\documentclass{article}
\usepackage{spconf,amsmath,graphicx,hyperref}

\usepackage{cite}
\usepackage{amsmath,amssymb,amsfonts}
\usepackage{algorithmic}
\usepackage{graphicx}
\usepackage{textcomp}
\usepackage{url}
\PassOptionsToPackage{table}{xcolor}
\usepackage{xcolor}
\usepackage{tcolorbox} 
\usepackage{booktabs} 
\usepackage{bbding}
\usepackage{multirow}
\definecolor{deep_blue}{RGB}{18,27,143}
\definecolor{deep_green}{RGB}{46,105,40}
\usepackage[hang,flushmargin]{footmisc}

\title{
FLEXI: Benchmarking Full-duplex Human-LLM Speech Interaction
}

\name{\begin{tabular}{c} Yuan Ge\textsuperscript{1$\dagger$}, Saihan Chen\textsuperscript{1$\dagger$}, Jingqi Xiao\textsuperscript{1}, Xiaoqian Liu\textsuperscript{1}, Tong Xiao\textsuperscript{1,2$\ddagger$}, \\ Yan Xiang\textsuperscript{3}, Zhengtao Yu\textsuperscript{3}, Jingbo Zhu\textsuperscript{1,2}
\end{tabular}
}
\address{\textsuperscript{1} School of Computer Science and Engineering, Northeastern University, Shenyang, China \\
    \textsuperscript{2} NiuTrans Research~~ 
    \textsuperscript{3} Kunming University of Science and Technology}

\begin{document}
\ninept
\maketitle

\begingroup\renewcommand\thefootnote{}\footnotetext{
$\dagger$ Equal contribution. $\ddagger$ Corresponding author. \\
\url{https://github.com/ChristineCHEN274/FLEXI}
}\endgroup

\begin{abstract}
Full-Duplex Speech-to-Speech Large Language Models (LLMs) are foundational to natural human-computer interaction, enabling real-time spoken dialogue systems. 
However, benchmarking and modeling these models remains a fundamental challenge. 
We introduce \texttt{FLEXI}, the first benchmark for full-duplex LLM–human spoken interaction that explicitly incorporates model interruption in emergency scenarios.
\texttt{FLEXI} systematically evaluates the latency, quality, and conversational effectiveness of real-time dialogue through six diverse human-LLM interaction scenarios, revealing significant gaps between open source and commercial models in emergency awareness, turn terminating, and interaction latency. Finally, we suggest that next token-pair prediction offers a promising path toward achieving truly seamless and human-like full-duplex interaction.
\end{abstract}
\begin{keywords}
Speech Dialogue Modeling, Human-LLM Interaction, Full-Duplex Dialogue, Turn-Taking
\end{keywords}

\section{Introduction}

The advent of large language models (LLMs) has revolutionized human-computer interaction (HCI), yet the ultimate goal remains a seamless, natural dialogue that mirrors human conversation \cite{clark1996using,luger2016like,ge2025sagelm}. The next frontier in this pursuit is full-duplex speech-to-speech (S2S) communication, where the forms of interaction extend far beyond turn-based S2S dialogue \cite{hurst2024gpt, kyutai2024moshi, xiong2024freeze}.

A key distinction of full-duplex speech conversation lies in its support for barge-in and natural turn-taking, which enable more flexible and human-like interactions. 
To delineate this continuum along the spectrum of intelligence and latency, we categorize duplex interaction into three levels: 
Level 1 mirrors face-to-face human conversation, characterized by immediacy and naturalness, typically with latency below $150$ ms. 
Level 2 resembles telephone communication—functional yet constrained by moderate delay or reduced expressiveness, requiring latency below $400$ ms. 
Level 3 corresponds to high-latency or low-intelligence exchanges, where responses appear mechanical or misaligned with human conversational norms.

\begin{figure}[tbp]
\centering
\centering
\includegraphics[width=0.86\columnwidth]{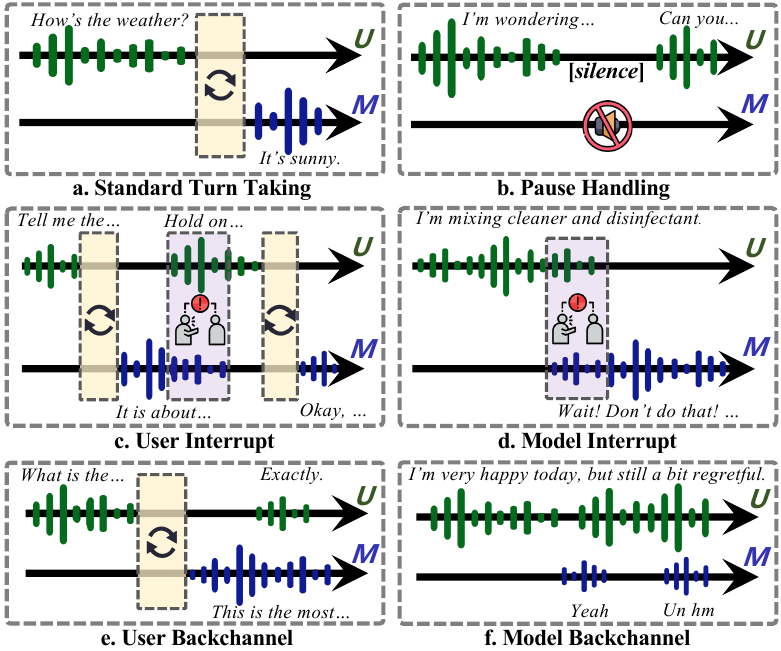}
\caption{Overview of six full-duplex scenarios in human-LLM interaction, while \textbf{\color{deep_green}{\textit{U}}} and \textbf{\color{deep_blue}{\textit{M}}} indicate \textbf{\color{deep_green}{User}} and \textbf{\color{deep_blue}{Model}}, respectively. }
\label{fig:case}
\end{figure}

Existing approaches to modeling full-duplex speech interaction can be broadly classified into two categories: 1) half-duplex S2S LLMs augmented with separate control modules \cite{fu2025vita, chen2025minmo, zhang2025llm, liao2025flexduo, lu2025duplexmamba}, and 2) end-to-end full-duplex S2S models \cite{kyutai2024moshi, xiong2024freeze, zhang-etal-2025-omniflatten, wang2024turn, yu2025salmonn, wangntpp, hu2025efficient}. 
The choice between these two methodologies remains inconclusive, as the former offers advantages in decoupling language modeling from duplex control, thereby reducing training overhead, while the latter provides superior low-latency performance.
However, we argue that this ambiguity arises from an incomplete understanding of the full-duplex interaction landscape. Recent modeling and benchmarking efforts have focused primarily on the isolated scenario of user interruptions, thereby overlooking the more comprehensive challenges inherent in streaming, complex, and diverse full-duplex interactions \cite{arora2025talking, lin2025full, lin2025evaluating}.

Motivated on this, we introduce \texttt{FLEXI}, a benchmark for \textbf{f}ull-dup\textbf{lex} LLM-human speech \textbf{i}nteraction, designed to evaluate the naturalness, seamlessness, and intelligence of S2S LLMs in challenging, complex, and diverse full-duplex scenarios.
All data are generated with advanced LLMs, synthesized into speech, concatenated into user queries, and manually validated.
Our contributions are: 
1) We introduce \texttt{FLEXI}, the first full-duplex benchmark encompassing six distinct human-LLM interaction scenarios which considers model interruption in emergency.
2) We conduct a comprehensive evaluation of the performance of open source duplex S2S LLMs, providing detailed insights into their capabilities and limitations.
3) We provide key insights into duplex model design, emphasizing that for real-time S2S LLMs to achieve Level 1 or Level 2 human-LLM interaction in the proposed benchmark scenarios, an end-to-end architecture based on next token-pair prediction paradigm is essential.

\section{Method}

\subsection{Problem Setup}

Compared to turn-based half-duplex dialogue models, prior work on full-duplex dialogue models introduces additional evaluation criteria \cite{wang2024turn, zhang-etal-2025-omniflatten, chen2025minmo, liao2025flexduo}: 
1) whether the model can produce a silence response to a user interruption within $k$ tokens; 2) whether the model can generate a non-silence response within $k$ tokens after the end of a semantically meaningful user speech segment; and 3) whether user backchannels disrupt the model’s ongoing output. However, these metrics offer only a fragmented assessment of model performance in full-duplex settings.

Building on prior work, Arora et al. \cite{arora2025talking} emphasizes the model’s ability to generate appropriate backchannels and its behavior during user pauses. However, the evaluation remains limited to Moshi and a cascaded system. 
Full-Duplex-Bench \cite{lin2025full} offers a more comprehensive evaluation of open source full-duplex dialogue models, but it overlooks the diversity of full-duplex interaction scenarios and the role of controllable modules in such systems.

To address these gaps, we propose \texttt{FLEXI}, a benchmark encompassing diverse full-duplex scenarios and including both open source and commercial systems. We adopt the formalization of turn-taking events following the methodology of dGSLM \cite{nguyen-etal-2023-generative} and Arora et al. \cite{arora2025talking}.
For simplicity, we consider a two speaker turn-taking event, where $k \in \{1,2\}$ represents two speakers, \textbf{\color{deep_green}{User}} and \textbf{\color{deep_blue}{Model}}. The state of speaker$_k$ at time $t$ is defined as:
\begin{equation}
    y_{t}^k = 
    \begin{cases}
        0, ~~~\text{speaker}_k~ \text{is speaking at time} ~t\\
        1, ~~~\text{speaker}_k~ \text{is \textit{not} speaking at time} ~t
    \end{cases}
\end{equation}

Further, we define the following concepts:

\begin{itemize}
\item \textit{Inter-Pausal Units} (\textit{IPU}) are continuous stretches of speech that are separated by a silence 
on each side. 
\begin{equation}
    \mathrm{IPU}^k = \{t \in [a,b]~ | ~y_{t}^k = 1\}
\end{equation}

\item \textit{Silences} are the durations that both speakers are not speaking, where $a^{sil}$ and $b^{sil}$ denote the start and end time.
\begin{equation}
    \mathrm{SIL} = \{t \in [a^{sil},b^{sil}]~ | ~y_{t}^1=0, ~y_{t}^2=0\}
\end{equation}

\item \textit{Pauses} are silences between IPUs of the same speaker.
Successive IPUs of the same speaker separated by pauses are grouped to form a \textit{turn}.
Formally, given $\mathcal{T}_j=\{i_1,i_2, ...,i_n\}$ is a series of IPU indices forming the $j$-th turn, and each adjacent pair of IPUs is separated by silence, we define:
\begin{equation}
\mathrm{Turn}_j^k = \bigcup_{i \in \mathcal{T}_j} \left( [a_i^k, b_i^k] \cup [b_i^k, a_{i+1}^k] \right)
\end{equation}

\item \textit{Gaps} are silences between IPUs of different speakers. Notably, gap is the signal preceding a turn-taking.
\item \textit{Overlaps} are times where there are IPUs for both speakers.
\begin{equation}
\mathrm{Overlap} = \{t \in [a^{o},b^{o}]~| ~y_{t}^1=1, ~y_{t}^2=1\}
\end{equation}

\item \textit{Backchannels} are overlaps such as `yeah' and `exactly', which the listener utters without taking the speaker’s turn to acknowledge the current speaker.
\textit{Interrupts} are overlaps between two speakers where both speakers are trying to take the turn. Formally, the IPUs of SPK2 interrupt SPK1 are:  
\begin{equation}
    \mathrm{IPU}^k = \{t \in [a^k,b^k]\} \quad \text{s.t.}~ a^1 < a^2 ~\text{and}~ a^2 < b^1
\end{equation}

\item An \textit{interrupt} is a successful interruption while $b^1 < b^2$ or an unsuccessful interruption while $b^1 > b^2$. In this paper interrupt indicates successful interrupt. 
\end{itemize}

\subsection{Evaluation Categories}

Building on this formalization, we categorize full-duplex user–LLM interactions into six key scenarios from the perspective of human–computer interaction, as illustrated in Fig. \ref{fig:case}.

\textbf{\textit{1) Standard turn-taking.}}~ This scenario mirrors traditional turn-based dialogue, where speakers alternate turns separated by short gaps. It evaluates whether the model can detect the end of a user's turn and respond promptly without undue delay.

\textit{Metric:} Following Lin et al. \cite{lin2025full}, Takeover Rate (TOR) is used to measure how often the model controls the conversation and dominates the turn. TOR is average across the dataset:
\begin{equation}
    \mathrm{TOR} = \frac{1}{N}\sum_{i=1}^n\mathrm{TO}_i, ~~\mathrm{TO} = 
    \begin{cases}
        0, ~~~\text{if SIL or BC}\\
        1, ~~~\text{otherwise}
    \end{cases}
\end{equation}

For successful takeovers, we calculate the latency between the end of the user speech and the start of the model response.

\textbf{\textit{2) Pause Handling.}}~ Humans naturally pause and hesitate between consecutive turns or within the same sentence. Additionally, user pauses much longer before or between utterances to think or organize their thoughts. Therefore, it is critical that the model does not takeover the turn prematurely during user pauses. Instead, it should decide whether to speak or yield based on semantic cues.
 
\textit{Metric:} 
We also use the TOR metric to measure whether the model takes over the turn within a time frame of the silence duration plus the model's average response latency.

\textbf{\textit{3) User Interrupt.}}~ A crucial aspect of full-duplex interaction, this scenario evaluates the model's ability to gracefully terminate its turn and attend to the new user input, as shown in Fig. \ref{fig:case}c.

\textit{Metric:} 
Performance is evaluated on three metrics. turn-taking Rate (TTR) measures the proportion of turns correctly terminated by the model, while latency quantifies the average time to termination. In addition, Topic Shift Score (TSS), rated by GPT-4o, assesses whether the model’s second-round response successfully shifts from the initial query to the user’s interrupting query.

\textbf{\textit{4) Model Interrupt.}}~ While often avoided in system design \cite{arora2025talking}, the ability for a model to proactively interrupt a user is vital in time-sensitive emergency contexts. Executing such interruptions presents a challenge in language understanding and latency control.

\textit{Metric:} We use TOR to assess if the model interrupts and an Emergency Detection Score (EDS) rated by GPT-4o to determine if the model correctly identifies the critical nature of the situation.

\textbf{\textit{5) User Backchannel.}}~ This scenario tests whether the model can ignore user backchannels (e.g., `yeah') without misunderstanding them as turn-taking attempts and prematurely yielding the floor.

\textit{Metric:} We use TTR to measure incorrect turn-yielding. An interruption is considered acceptable only if the model resumes its original utterance within 1 second, measured by a coherence metric.

\textbf{\textit{6) Model Backchannel.}}~ Although less emphasized in HCI, appropriate model-generated backchannels can signal active listening and enhance naturalness. As shown in Fig. \ref{fig:case}f, this scenario evaluates the model's ability to produce socially appropriate backchannels.

\textit{Metric:} We use TOR and Backchannel Rate (BCR) to quantify model behavior. To assess appropriateness, we use GPT-4o to annotate reference backchannel positions and compute the Jensen-Shannon Divergence (JSD) between the model's predicted backchannel distribution ($P$) and the reference distribution ($Q$). JSD scores range from 0 (perfect alignment) to 1 (complete divergence). 
\begin{equation}
\text{JSD}(P || Q) = \frac{1}{2} \sum_{i} P(i) \log \frac{P(i)}{M(i)} + \frac{1}{2} \sum_{i} Q(i) \log \frac{Q(i)}{M(i)}
\end{equation}
where \( M(i) = \frac{1}{2}(P(i) + Q(i)) \) and i index the time windows.

\section{Experiments and Analysis}

\subsection{Experimental Setups}

\begin{figure}[tbp]
\centering
\centering
\includegraphics[width=0.96\columnwidth]{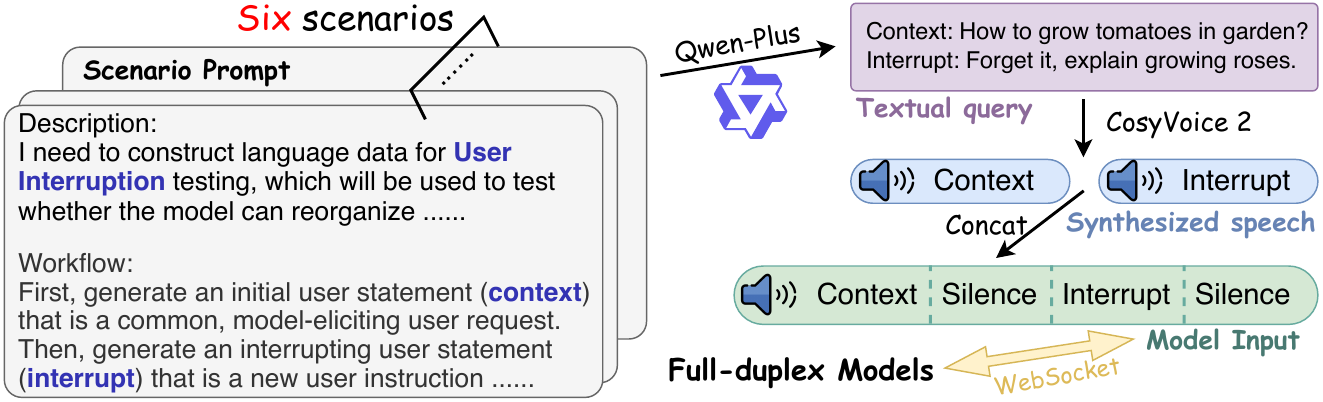}
\caption{Overview of data construction and evaluation pipeline. }
\label{fig:pipeline}
\end{figure}

\noindent \textbf{\textit{Dataset.}}~
Owing to the scarcity of real full-duplex conversational data, we follow Lin et al. \cite{lin2025full} and evaluate the interaction performance of S2S full-duplex LLMs using synthetic data. As shown in Fig. \ref{fig:pipeline}, user queries across six scenarios are generated with Qwen-plus\footnote{\url{https://www.alibabacloud.com/help/en/model-studio/what-is-qwen-llm}} and  synthesized into speech with CosyVoice2 \cite{du2024cosyvoice}. To approximate natural interaction, we insert appropriate silences after user backchannels, interruptions, and query completions to allow the model to produce full responses. Code and dataset are available. 

\noindent \textbf{\textit{Models.}}~ 
We summarize existing S2S full-duplex models in Table \ref{tab:Models}, including both transparent and commercial systems. For evaluation, we tested open source transparent models and only the commercial models Gemini due to budget constraints. Furthermore, the real-time full-duplex interaction between human and LLM is simulated utilizing WebSocket\footnote{\url{https://ai.google.dev/gemini-api/docs/live}}, more details are available at our project page.

\begin{table}[!h]
    \centering
    \small
    \begin{tabular}{lccc}
        \toprule
        Model & E2E FD & Open Source & API Access \\ 
        \midrule
        \multicolumn{4}{l}{\cellcolor{black!10} \textbf{\textit{I. Transparent Models}}} \\
        Moshi\cite{kyutai2024moshi} & \Checkmark & \Checkmark & - \\ 
        OmniFlatten\cite{zhang-etal-2025-omniflatten} & \Checkmark & \XSolidBrush & - \\ 
        Salmonn-omni\cite{yu2025salmonn} & \Checkmark & \XSolidBrush & - \\ 
        NTPP\cite{wangntpp} & \Checkmark & \XSolidBrush & - \\ 
        SALM-duplex\cite{hu2025efficient} & \Checkmark & \XSolidBrush & - \\ 
        Freeze-Omni\cite{xiong2024freeze} & \XSolidBrush & \Checkmark & - \\
        Vita1.5\cite{fu2025vita} & \XSolidBrush & \Checkmark & - \\         
        MinMo\cite{chen2025minmo} & \XSolidBrush & \XSolidBrush & - \\ 
        SDS\cite{zhang2025llm}& \XSolidBrush & \XSolidBrush & - \\ 
        FlexDuo\cite{liao2025flexduo} & \XSolidBrush & \XSolidBrush & - \\ 
        \midrule
        \multicolumn{4}{l}{\cellcolor{black!10} \textbf{\textit{II. Closed-source Commercial Models}}} \\
        GPT-4o-realtime & - & \XSolidBrush & \Checkmark \\ 
        Gemini-2.5-flash-live & - & \XSolidBrush & \Checkmark \\ 
        Nova Sonic & - & \XSolidBrush & \Checkmark \\ 
        DouBao & - & \XSolidBrush & \XSolidBrush \\ 
        \bottomrule
    \end{tabular}
    \caption{Overview of full-duplex speech language models. `—' denotes properties that are either unspecified or transparent in models that do not require API access. `E2E FD' indicates whether a model supports end-to-end full-duplex interaction without auxiliary control modules. `Open Source' specifies whether the complete S2S full-duplex dialogue pipeline is publicly released, and `API Access' denotes whether commercial models provide API availability.}
    \label{tab:Models}
\end{table}

\subsection{Experimental Results}

\noindent \textbf{Turn-Taking and Pause Handling.}~ Table \ref{tab:result1} shows that models generally take over the turn with high probability when prompted with a user query, with end-to-end systems exhibiting lower latency. However, we observe that the short delays of Moshi are largely attributable to an aggressive turn-taking strategy: as jump-in rate (JIR), Moshi may initiate a response before the user input is complete, a phenomenon rarely seen in commercial systems. For pause handling, all models except Moshi achieve around 90\% TOR, reflecting the same aggressive strategy whereby the system prematurely takes over during user pauses, leading to poor conversational experience. These findings highlight a trade-off between standard turn-taking and pause handling, determined by whether the architecture adopts aggressive turn-taking strategy.

\begin{table}[t]
    \centering
    \begin{tabular}{lcccc}
    \toprule
        \multirow{2}{*}{Model} & \multicolumn{3}{c}{turn-taking} & Pause Handing \\ 
        \cline{2-5}
        \specialrule{0em}{1pt}{1pt}
        & TOR($\uparrow$) & Latency & JIR($\downarrow$) & TOR($\downarrow$) \\ 
        \midrule
        Moshi & 0.995 & \textbf{0.696} & 0.785 & \textbf{0.550}  \\ 
        Freeze-omni & \textbf{1.000} & 1.147 & 0.120 & 0.937 \\ 
        Vita1.5 & \textbf{1.000} & 4.878 & \textbf{0.000} & 0.916 \\
        Gemini & \textbf{1.000} & 1.736 & \textbf{0.000} & 0.835 \\ 
        % Sonic & 0.915 & 2.765 & 0.005 & 0.480 \\ 
        \bottomrule
    \end{tabular}
    \caption{Experimental results on turn-taking and pause handing scenarios. TOR indicates takeover rate and jump-in rate (JIR) indicates the model takeover the turn before user turn completed.}
    \label{tab:result1}
\end{table}

% \begin{table*}[!ht]
%     \centering
%     \begin{tabular}{lccccccccc}
%     \toprule
%         \multirow{2}{*}{Model} & \multicolumn{3}{c}{User Interrupt} & \multicolumn{2}{c}{Model Interrupt} & \multicolumn{2}{c}{User Backachannel} & \multicolumn{2}{c}{Model Backchannel}   \\
%         \cline{2-10}
%         \specialrule{0em}{1pt}{1pt}
%         & TOR($\uparrow$) & Latency($\downarrow$) & TSS($\uparrow$) & IR($\uparrow$) & EDS($\uparrow$) & IR($\downarrow$) & coherence & TOR($\downarrow$)& JSD($\downarrow$)  \\ 
%         \midrule
%         Moshi & 0.970 & 0.258 & 2.48 & 0.72 & 0.77 & 0.04 & 0/8 & 0.553 & 0.932 \\ 
%         Freeze-omni & 1.000 & 1.101 & 2.19 & 0.56 & 1.84 & 0.13 & 0/24 & 0.095 & 1.000 \\ 
%         Gemini & 0.995 & 0.026 & 1.28 & 0.32 & 2.72 & 0.03 & 0/6 & 0.141 & 0.959 \\ 
%         Sonic & 0.944 & 3.521 & 1.25 & 0.01 & 3.12 & 0.00 & 0/0 & 0.015 & 0.992 \\
%         \bottomrule
%     \end{tabular}
%     \label{tab:result2}
%     \caption{Experimental results about user interrupt, model interrupt, user backachannel and model backchannel.}
% \end{table*}

\noindent \textbf{User and Model Interrupt.}~ As shown in Table \ref{tab:result2}, commercial model significantly outperforms open source models in managing interruptions. Gemini effectively handles interruptions by terminating its current output and generating a response directly addressing the user's query, as reflected by the high Topic Shift Score (TSS) of GPT-4o. 
Additionally, the objective of model-initiated interruption is to identify and react to emergency user states. Although open source systems interrupt more frequently, their higher rates stem primarily from flawed turn-taking rather than successful emergency detection. Consequently, their Emergency Detection Scores (EDS) are low, as their responses fail to acknowledge the emergency context. In contrast, commercial models demonstrate the capacity to recognize emergencies within their responses, but their interaction strategies appear to favor polite listening over assertive interruption.

\begin{table}[t]
    \centering
    \begin{tabular}{lccccc}
    \toprule
        \multirow{2}{*}{Model} & \multicolumn{3}{c}{User Interrupt} & \multicolumn{2}{c}{Model Interrupt} \\
        \cline{2-6}
        \specialrule{0em}{1pt}{1pt}
        & TTR($\uparrow$) & Latency & TSS($\uparrow$) & TOR($\uparrow$) & EDS($\uparrow$) \\ 
        \midrule
        Moshi & 0.462 & 2.660 & 2.48 & 0.390 & 0.81 \\ 
        Freeze-o & 0.539 & \textbf{1.751} & 2.90 & \textbf{0.535} & 1.84 \\ 
        Vita1.5 & 0.226 & 4.016 & 0.75 & 0.000 & 1.72 \\ 
        Gemini & \textbf{0.935} & 1.766 & \textbf{4.51} & 0.000 & \textbf{2.19} \\ 
        % Sonic & 0.944 & 3.521 & 1.25 & 0.01 & 3.12 \\
        \bottomrule
    \end{tabular}
    \caption{Experimental results on user and model interrupt scenarios. TTR indicates the rate of model turn termination. Additionally, Topic Shift Score (TSS) and Emergency Detection Score (EDS) is scored by GPT-4o ranging from 0 to 5, indicating whether model response related to the interruption and emergency case, respectively.}
    \label{tab:result2}
\end{table}

\begin{figure*}[htbp]
\centering
\centering
\includegraphics[width=1.93\columnwidth]{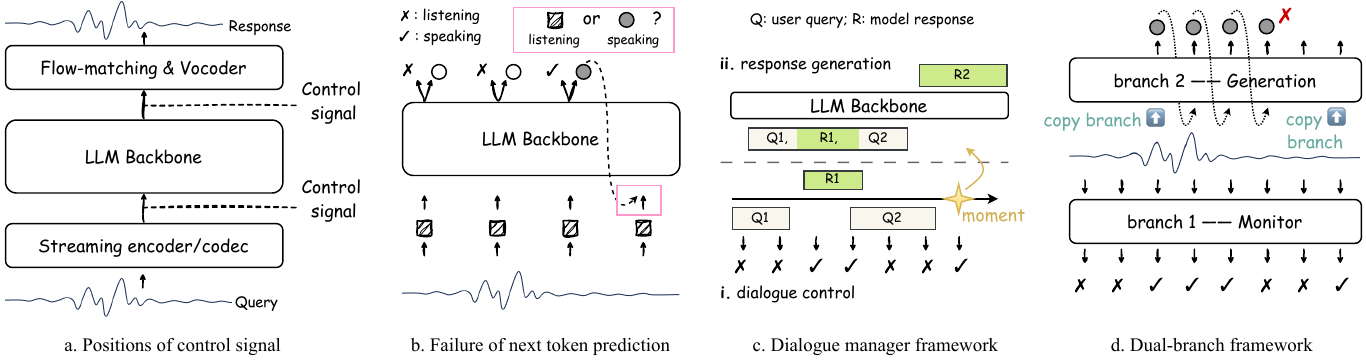}
\caption{Discussion: Key Challenges in Full-duplex Modeling and Primacy of End-to-End Full-Duplex Modeling.}
\label{fig:discussion}
\end{figure*}

\noindent \textbf{User and Model Backchannel.}~ Table \ref{tab:result3} further shows that user backchannels almost never interrupt model output, suggesting that system responses are not disrupted by human acknowledgments. Notably, in 4 out of the 6 instances where Gemini was interrupted, it continued with the previous topic. Furthermore, all models demonstrated weak backchanneling ability. Instead, they frequently misinterpret user speech as a signal to take over, and none of the systems aligned their backchannel timing with GPT-4o’s appropriate annotations. Interestingly, we observed several strong backchannel cases in the model-interrupt scenario, suggesting that models possess the capability but lack accurate timing in deploying backchannels.

\begin{table}[t]
    \centering
    \begin{tabular}{lccccc}
    \toprule
        \multirow{2}{*}{Model} & \multicolumn{2}{c}{User Backachannel} & \multicolumn{3}{c}{Model Backchannel}   \\
        \cline{2-6}
        \specialrule{0em}{1pt}{1pt}
        & TTR($\downarrow$) & Coh. & TOR($\downarrow$)& BCR($\uparrow$) & JSD($\downarrow$)  \\ 
        \midrule
        Moshi & 0.020 & 1/4 & 0.578 & \textbf{0.031} & \textbf{0.947} \\ 
        Freeze-o & 0.086 & 8/17 & 0.608 & 0.003 & 0.986 \\ 
        Vita1.5 & \textbf{0.000} & 0/0 & 0.578 & 0.002 & 0.986 \\ 
        Gemini & 0.031 & 4/6 & \textbf{0.045} & 0.012 & 0.981 \\ 
        % Sonic & 0.00 & 0/0 & 0.035 & 0.003 & 0.990 \\
        \bottomrule
    \end{tabular}
    \caption{Experimental results on user model backchannel scenarios. Coherence (Coh.) indicates whether models resume the previous topic after being interrupted by a backchannel. Backchannel rate (BCR) accesses whether the model exhibits backchannel and Jensen–Shannon Divergence (JSD) quantifies the consistency between model-predicted backchannels and GPT-4o annotations.}
    \label{tab:result3}
\end{table}

% \begin{table*}[!ht]
%     \centering
%     \small
%     \begin{tabular}{lccccccccccccc}
%     \toprule
%         \multirow{2}{*}{Model}& \multicolumn{3}{c}{turn-taking} & Pause & \multicolumn{3}{c}{User Interrupt} & \multicolumn{2}{c}{Model Interrupt} & \multicolumn{2}{c}{User Backachannel} & \multicolumn{2}{c}{Model Bachchannel}   \\ \cline{2-14}
%         & TOR & latency & JIR & TOR & TOR & latency & JIR & Num & UDS & IR & coherence & TOR & JSD  \\ 
%         \midrule
%         Moshi & 99.5 & 0.696 & 78.5 & 54.0 & 97.0 & 0.258 & 2.479 & 144 & 0.77 & 4.0 & 0/8 & 55.28 & 93.17 \\ 
%         freeze-o & 100 & 1.141 & 12.0 & 49.0 & 100 & 1.101 & 2.19 & 111 & 1.84 & 12.57 & 0/24 & 9.52 & 1.0 \\ 
%         gemini & 82.0 & 3.11 & 1.0 & 47.0 & 99.5 & 0.0257 & 1.276 & 64 & 2.72 & 3.06 & 0/6 & 14.07 & 95.9 \\ 
%         sonic & 91.5 & 2.765 & 0.5 & 48.5 & 94.42 & 3.5208 & 1.252 & 1 & 3.12 & 0 & 0/0 & 1.51 & 99.23 \\ 
%         \bottomrule
%     \end{tabular}
% \end{table*}

\section{Discussion: Full-duplex Modeling}

Existing modeling paradigms for full-duplex speech interaction fall into two categories: 1) half-duplex S2S LLMs with separate control modules, and 2) end-to-end full-duplex S2S models. The six scenarios in \texttt{FLEXI} underscore the necessity of the end-to-end approach.

\noindent \textbf{Streaming Comprehension.}~ Full-duplex S2S LLM necessitates a streaming encoder or codec module \cite{ijcai2024p0900, ji2024wavchat}. A well-trained module can alleviate the complexities of pre-training the full-duplex model.

\noindent \textbf{Turn-Taking Determination.}~ Deciding when to speak and when to yield presents a critical challenge in dialogue state management. As illustrated in Fig. \ref{fig:discussion}a, control signals can be derived from embedded representations or hidden states. However, scenarios requiring model interruption demand that the control module identify emergency situations, which necessitates leveraging the linguistic priors and reasoning processes encapsulated within the LLM parameters.

\noindent \textbf{Listening While Speaking.}~ A fundamental conflict exists between this core requirement of full-duplex modeling and the next-token prediction (NTP) paradigm. As depicted in Fig. \ref{fig:discussion}b, once the model begins response generation, the autoregressive nature of NTP precludes the processing of subsequent streaming audio input. One solution (Fig. \ref{fig:discussion}c) employs an external control module that postpones response generation until the user's query is complete; however, this two-pass comprehension process nearly doubles latency. 
An alternative dual-branch approach (Fig. \ref{fig:discussion}d) trades computational space for time: when the system decides to speak, it forks a parallel branch to continue generating control signals while the primary branch generates the response. The critical limitation here is the inability to model overlap, such as when a generated response collides with a user interruption, which leads to incomplete dialogue history.

\noindent \textbf{Latency.}~ Ultimately, full-duplex modeling requires navigating a critical trade-off between conversational intelligence and latency. The acceptable threshold for human conversational latency is under $400$ ms, a stringent constraint that challenges non-end-to-end pipeline designs. For example, VITA's latency is around $4$ s, while commercial models such as GPT and Gemini are around $1$ s.
\footnote{\url{https://artificialanalysis.ai/models/speech-to-speech}}).

In summary, the combined challenges of diverse interaction scenarios, conversational intelligence, and latency constraints collectively motivate a native, dual stream full duplex architecture. By employing next token-pair prediction, such a model can achieve concurrent listening and speaking, thereby preserving conversational intelligence while significantly reducing interaction latency.
\vspace{-2pt}
\begin{tcolorbox}[colback=blue!5!white, colframe=blue!75!black, title=\textbf{Take away finding:}]
\textbf{Primacy of End-to-End Full-Duplex Modeling by Next Token-pair Prediction.}  The combined demands for conversational intelligence, concurrent listening and speaking, and low latency challenge the viability of the technical approach that pairs half-duplex models with dialogue management modules. In contrast, an end-to-end full duplex conversational modeling paradigm, based on next token-pair prediction, shows significantly greater promise.
\end{tcolorbox}
\vspace{-2pt}

\section{Conclusion}
We introduced \texttt{FLEXI}, a benchmark for full-duplex spoken interaction featuring six diverse human-LLM scenarios. Our comprehensive evaluation uncovered key limitations in current open source and commercial models, such as high latency, aggressive turn-taking, ineffective turn-yielding upon user interruption, a lack of proactive interruption in emergencies, and minimal system backchanneling. These results provide a roadmap for guiding the development of future full-duplex systems and data collection pipelines. Furthermore, we argue that next token-pair prediction offers a promising methodology for enabling more seamless and human-like interaction.

\bibliographystyle{IEEEbib}
\bibliography{strings,refs}

\end{document}